\documentclass{article}

\PassOptionsToPackage{hyphens}{url}

\usepackage{arxiv_ml_institute}

\usepackage[utf8]{inputenc} \usepackage[T1]{fontenc}    \usepackage{hyperref}       \usepackage{url}            \usepackage{booktabs}       \usepackage{amsfonts}       \usepackage{nicefrac}       \usepackage{microtype}      \usepackage{xcolor}         \usepackage{graphicx}       

\usepackage{algorithm}
\usepackage{algorithmic}

\usepackage{ml_defs}

\usepackage[textsize=tiny]{todonotes}

\DeclareMathOperator*{\drawnfrom}{\sim}
\newcommand{\methodname}{RREDCoT}
\newcommand{\methoddisambig}{Reward REDistribution for Chain of Thoughts}

\title{\methodname{}: Segment-Level Reward \\ Redistribution for Reasoning Models}

\author{
	Mykyta Ielanskyi$^{1}$, \
	Kajetan Schweighofer$^{2}$\thanks{Work done while at Johannes Kepler University Linz.} , \
	Lukas Aichberger$^{1}$, \
	Sepp Hochreiter$^{1,3}$ \\
	$^1$~ELLIS Unit Linz and LIT AI Lab, Institute for Machine Learning, \\ 
	\ \ \ Johannes Kepler University Linz, Austria \\
	$^2$ Cognizant AI Lab, San Francisco, USA \\
	$^3$~NXAI GmbH, Linz, Austria\\
	\ \ \texttt{\{ielanskyi, schweighofer, aichberger, hochreit\}@ml.jku.at}
}

\begin{document}
	
	\maketitle
	
	\maketitle
	
	\maketitle
	
	\begin{abstract}
		
		Recent advancements in reasoning language models have been driven by Reinforcement Learning (RL) fine-tuning.
		Most often, these rely on the Group Relative Policy Optimization (GRPO) algorithm or modifications thereof to steer the models to produce Chain-of-Thought (CoT) traces.
		The final answer can only be verified, and the reward assigned, after the CoT trace is complete, making it a delayed reward problem.
		GRPO and its modifications correspond to Monte Carlo methods in standard RL, which are known to suffer from high variance. 
		A possible solution to this problem is the redistribution of rewards through credit assignment, where segments of the CoT trace that are important for arriving at the desirable solution are emphasized by assigning a higher reward.
		While Monte Carlo sampling can be used to provide an unbiased estimate of intermediate state values, its computational overhead makes it unsuitable for train-time credit assignment in long contexts at high granularity.
		We introduce \methodname{} (\methoddisambig{}), which utilizes the model itself to approximate the optimal reward redistribution without additional generation.
		We investigate the advantages of our method compared to MC sampling and several attribution methods.
		We further analyze several aspects relevant to the construction of the redistribution such as segmentation of CoT traces and state value estimation. 
	\end{abstract}
	
	\section{Introduction}
	
	As established modes of language model pretraining hit the boundaries of data availability, the focus shifts further towards maximizing the utility extracted from the pretrained models in subsequent ``post-training'' stages \citep{Team:2025}.
	Techniques such as Reinforcement Learning from Human Feedback (RLHF) \citep{Ouyang:2022} have emerged, relying on a reward signal from possibly vague human preference data.
	These early attempts to improve the quality of language models were based on the Proximal Policy Optimization (PPO) algorithm \citep{Schulman:17}.
	
	Today, Group Relative Policy Optimization (GRPO) \citep{Shao:2024} has become the most popular choice for RL-based fine-tuning, because it circumvents the necessity of simultaneously learning an advantage function.
	This is combined with RL with Verifiable Rewards (RLVR) paradigm, in which model generates a Chain-of-Thought (CoT) before producing the final result, with training often consisting of online tuning followed by distillation on the generated reasoning traces \citep{DeepSeek-AI:2025}.
	However, CoT generation poses several challenges: models tend to produce excessively long traces, and answer extraction and reward estimation can yield disparate performance assessments and noisy training signals (\citet{Shao:2025}, \citet{Chandak:2025}).
	
	One prominent issue that RL fine-tuning of reasoning language models still faces is the lack of fine-grained reward signals. 
	In RLVR, the reward for the entire episode is assigned only at the very end of a generated Chain-of-Thought (CoT) and distributed uniformly over the full trajectory, which provides no direct supervision for individual reasoning steps. 
	In on-policy reasoning fine-tuning, this problem has been approached from several directions. 
	Some works propose step-by-step analysis of CoT traces using judge models as a means of credit assignment \citep{Xie:2025a, Ou:2025, Jayalath:2025}, while others attempt to extract intermediate utility directly from model statistics \citep{Li:2026}. 
	Monte Carlo sampling (MCS) based state value estimation methods have been shown to provide effective intermediate value estimates \citep{Kazemnejad:2025, Guo:2025} at the cost of considerable additional token generation.
	
	With \methodname{}, we adapt the core principles of RUDDER \citep{Arjona-Medina:2019} to the specific structure of the CoT generation MDP and derive a tractable approximation of the optimal reward redistribution that requires neither additional models nor extra generation steps.
	Unlike the original RUDDER formulation, which employs an LSTM \citep{Hochreiter:1997} for return decomposition, we utilize the generating language model itself. 
	This exploits the properties of autoregressive sequence generation together with a novel fast value function estimator inspired by Bayesian methods in natural language generation \citep{Malinin:2020, Aichberger:2024a}.
	
	\begin{figure*}[t]
		\centering
		\includegraphics[width=0.95\linewidth]{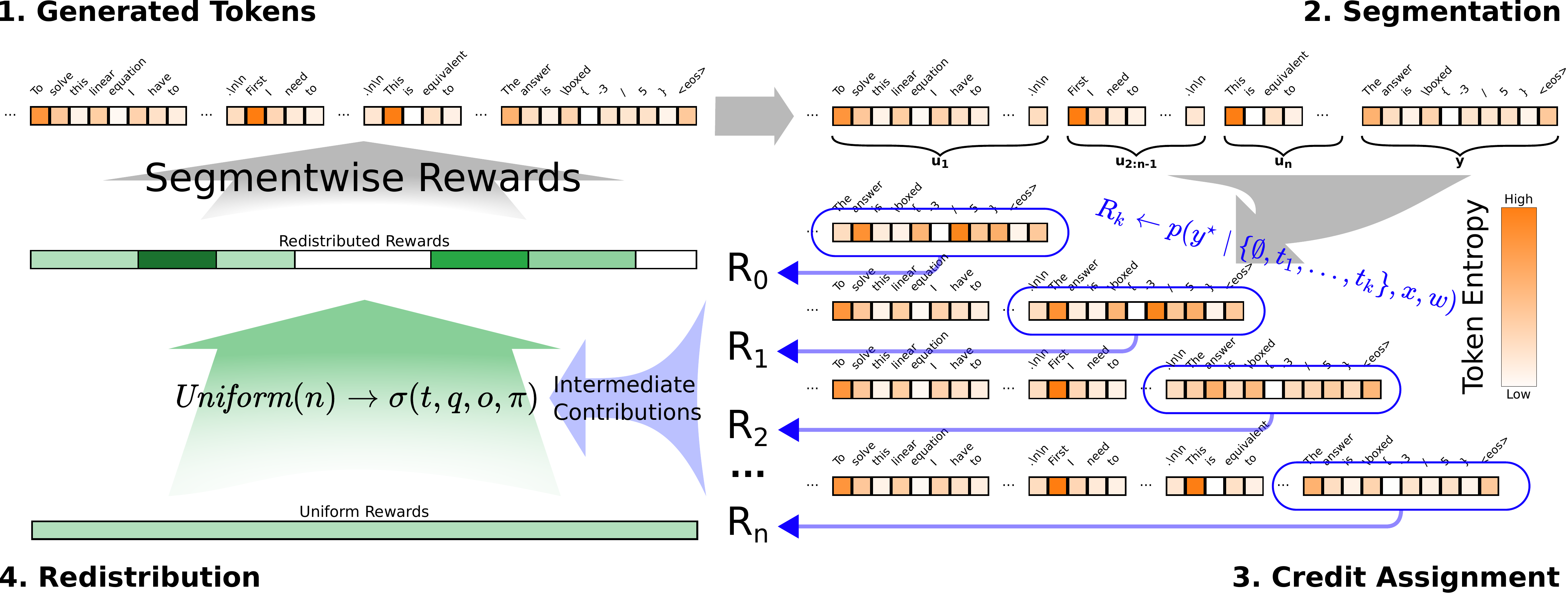}
		\caption{Overview of the \methodname{} algorithm for reward redistribution. To compute a reward redistribution $\sigma$: (1) generate a CoT trace, (2) segment the trace, (3) compute each segment’s immediate reward, and (4) use these rewards to derive the final redistribution.}
		\label{fig:method-overview-fig-1}
	\end{figure*}
	
	Our contributions are as follows:
	\begin{itemize}     \item We introduce \methodname{} - a tractable credit assignment and reward redistribution algorithm for CoT traces that does not require additional models.
		\item We devise an entropy-based segmentation strategy for the CoT traces that is specifically tailored for \methodname{}.
		\item We analyze the relation between our estimator, MC sampling and several attribution methods.\end{itemize}
	
	\section{CoT Generation as a Reinforcement Learning Problem}
	
	We consider a Markov Decision Process $\cP = (\cS, \cA, \cR, p, \gamma)$ where $\cS$ is a set of states, $\cA$ the set of actions, and $\cR$ the reward space. 
	A state at step $t$ is given by $s_{t} = ( \Bx, \Bu_1, \dots, \Bu_{t-1} )$ where $\Bx$ is the original query and each $\Bu_t$ is a generated CoT \emph{segment}.
	The language model is then the policy with a discrete categorical action space, i.e., the space of possible next tokens $a$, also called vocabulary.
	
	Segments are contiguous and non-overlapping sequences of generated tokens that together span the entire sequence.
	In the corner cases, the whole generated text would be split into segments consisting of individual tokens or kept together as a large single segment.
	This depends on \emph{segmentation strategy} discussed further in Sec.~\ref{sec:segment-treatment}.
	Importantly, as the state is defined by $\Bs_{t} = ( \Bx, \Bu_1, \dots, \Bu_{t-1} )$, the dynamics function is deterministic when conditioned on the next segment, i.e., $p(\Bs_{t+1} \mid \Bs_t,\Bu_t) = 1$.
	Thus, there is only stochasticity through the policy (the LLM).
	In the subsequent equations, whenever the transitions (e.g. $p(\Bs_{t+1} \mid \Bs_{t})$) are listed without specifying the parameters, $\Bw$ of the generating model are implied. 
	
	\paragraph{Reward Modeling for CoT Outputs.}
	
	The reward model of text generation used for evaluation is as per Eq.~\eqref{eq:reward_equation_cot}.
	We call $\zeta : \By \mapsto \left[0, 1\right]$ a utility function.
	It maps from the answer space to a real value, which is the inverse of the cost of generating the final output sequence $\By$. 
	In the simplest case, $\zeta$ is defined as the correctness of the answer.
	Often it is a combination of several different cost functions, such as adherence to output format or the overall length of $(\Bu, \By)$, with improperly formatted or overly long sequences being associated with higher cost.
	The return is then the expected inverse risk, conditioned on the CoT trace $\Bu$.
	\begin{align}
	R_{u} = \EXP_{\By \drawnfrom p(\By \mid \Bu, \Bx, \Bw)}[\zeta(\By)]
	\label{eq:reward_equation_cot}
	\end{align}
	The CoT trace $\Bu$ does not affect the reward assigned to the final action $\By$ directly. Instead, it affects the reward by changing the conditional probability distribution of $\By$.
	The VR estimator uses Monte Carlo integration to estimate $R_{\By}$ with most prior work only sampling a single output together with the original trace (Appx.Eq.~\eqref{appx:eq:ry_with_verifiers_mc_estimate}).
	Several recent works propose using Probability Reward (PR) instead \citep{Zhou:2025, Yu:2025}.
	PR reward estimators use importance sampling (Appx.Eq.~\eqref{appx:eq:ry_probability_reward_importance_sampling}) to reduce variance. 
	This way, the ability of a model to serve as a dynamics function (i.e., world model) is leveraged.
	It is thus important that the model is reasonably well calibrated to assess the degree of causality between the CoT trace and the answer.
	PR estimator provides a denser signal but may suffer from increased noise as was pointed out in \citet{Yu:2025a}.
	
	\paragraph{CoT Generation MDP and its Bellman Equation.}\label{sec:cot-generation-mdp}
	
	\begin{figure}
		\centering
		\includegraphics[width=0.45\linewidth]{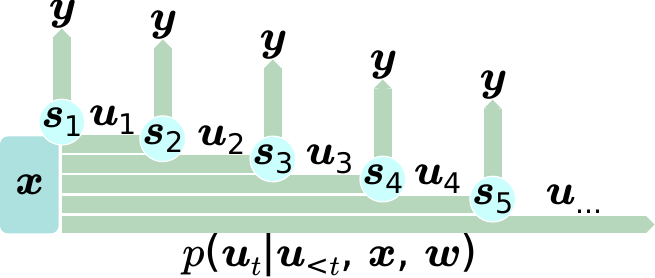}
		\hspace{0.05\linewidth}
		\includegraphics[width=0.45\linewidth]{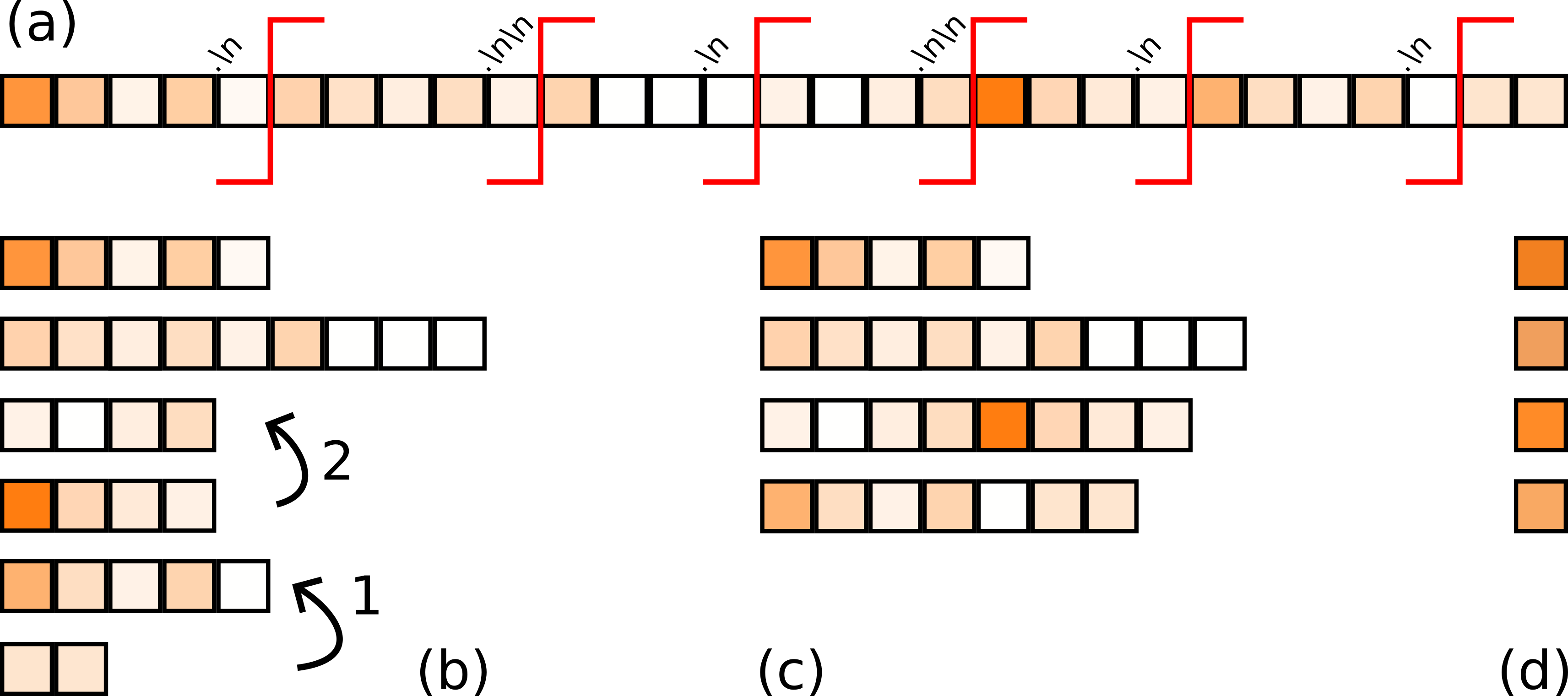}
		\caption{\textbf{(Right)} The MDP structure of autoregressive CoT generation. 
			At each point, either another segment $\Bu_{t+1}$ or the answer $\By$ can be generated.
			The state $\Bs_{t}$ at each point consists of the entire sequence generated so far and the original input: $\{ \Bx, \Bu_1 \dots \Bu_{t-1} \}$. 
			\textbf{(Left)} Hybrid segmentation approach. \emph{(a)} keyword segmentation; \emph{(b)} iterative merging of the lowest total entropy pairs; \emph{(c)} merging stops when criteria is met (number of segments); \emph{(d)} the segment entropies are homogenized.
		}
		\label{fig:cot-mdp-and-segmentation}
	\end{figure}
	A peculiar feature of the particular MDP at hand is that the action space $\cA$ is partitioned into $\{\cY, \cU\}$ subspaces.
	The $\cY$ subspace contains all the actions that result in the beginning of the final output, while the $\cU$ subspace contains all of the actions that do not trigger the output and instead continue the generation of the CoT trace. 
	
	In the commonly used models, the first subset of actions is initiated when end of thinking token (EOT) is produced, i.e. \textless/think\textgreater.
	Upon selecting the EOT token, the model can then produce the answer for which the reward can be evaluated as per Eq.~\eqref{eq:reward_equation_cot}.
	Therefore, we can rewrite the Bellman equation for such MDP in the following fashion:
	\begin{align}
	v_{\Bw} (\Bs_{t}) = & \sum_{a \in \cY} p(a \mid \Bs_t) \; \zeta(\Bs_{t+1})  + \sum_{a \in \cU} p(a \mid \Bs_{t}) \ \gamma \ v_{\Bw}(\Bs_{t+1})  \nonumber \\
	= & \: \EXP_{\By \drawnfrom p(\By \mid \Bs_{t})} [ \zeta(\By) ]+ \EXP_{\Bu_t \drawnfrom p(\Bu_t \mid \Bs_{t})} [ v_{\Bw}(\Bs_{t+1}) ] \label{eq:cot-mdp-bellman}
	\end{align} 
	
	In Eq.~\eqref{eq:cot-mdp-bellman}, we use the knowledge of the CoT MDP structure to split the value function on the action space $\cA$ into two parts: one that corresponds to the model selecting an answer and the other that corresponds to the model selecting to continue CoT generation with fragment $\Bu_t$.
	With CoT generation, the model only gets rewarded for generating useful output in the end. 
	The immediate reward for generating another thought is zero.
	CoT traces from commonly used models can reach thousands of tokens in length \citep{Muennighoff:2025, DeepSeek-AI:2025}, resulting in highly delayed rewards.
	We can further derive the corresponding action-value function $q^{\Bw} (\Bs_{t}, a)$:
	\begin{align}
	q^{\Bw} (\Bs_{t}, a) = 
	\begin{cases} \EXP_{a \drawnfrom p(a \mid \Bs_t, \Bx, \Bw)} [ \zeta(\Bs_{t}) ] & \text{if } a \in \cY \\ 
	v_{\Bw}(\Bs_{t+1}) & \text{if } a \notin \cY 
	\end{cases} \label{eq:cot-mdp-action-value}
	\end{align} 
	
	Alternatively, the value term of the CoT MDP can be interpreted as an expectation under the predictive distribution of the reasoning language model:
	\begin{align}
	v_{\Bw} (\Bs_{t}) = & \; \EXP_{\By, \Bu \drawnfrom p(\By, \Bu \mid \Bx, \Bw)} \bigg[ \zeta(\By) \bigg] \nonumber \\
	= & \; \sum_{\By \in \cY, \Bu \in \cU} p(\By, \Bu \mid \Bx, \Bw) \cdot \zeta(\By) \nonumber \\
	= & \; \sum_{\By \in \cY, \Bu \in \cU} p(\By \mid \Bu, \Bx, \Bw) p(\Bu \mid \Bx, \Bw) \cdot \zeta(\By) \label{eq:cot-mdp-bellman-bayesian}
	\end{align}
	In this formulation, the reward function $\zeta$ is integrated over the space of $\BY \times \BU$ that consists of the final outputs $\By$ and chains of thought $\Bu$. 
	What enables such a transformation is that (a) the dynamics function is trivial; (b) the intermediate rewards for transitions within $\Bu$ are zero; (c) $\gamma$ is normally ignored and set to one in the CoT setting.
	This view enables us to use some of the insights from Bayesian methods applied to LLMs to estimate the value term. 
	
	\section{\methodname{}}
	
	\renewcommand{\algorithmicensure}{\textbf{Supplies:}}
	\renewcommand{\algorithmicrequire}{\textbf{Requires:}}
	\begin{algorithm}[t!]
		\caption{ Approximately Optimal Reward Redistribution for CoT traces. }
		\label{alg:cotrudder_redist}
		\begin{algorithmic}[1]
			\ENSURE Provides the optimal redistribution of Conditional Code Length 
			\REQUIRE CoT trace of $T$ segments $( \Bu_1, \dots, \Bu_T )$, reference answer $\By^\star$, reference solution $\Bu^\star$, original input sequence $\Bx$, predictive model with parameters $\Bw$ used for autoregression. 
			\FOR{ $t=0$ to $T$ }
			\STATE $\hat{R}_t \leftarrow p( \By^\star \mid (\emptyset, \Bu_1, \dots, \Bu_t ),\Bx, \Bw)$ \qquad \textcolor{gray}{// the answer probs are evaluated in prefill mode}
			\STATE $\hat{v}_t \leftarrow p( \Bu^\star \mid (\emptyset, \Bu_1, \dots, \Bu_t ),\Bx, \Bw)$ \qquad \textcolor{gray}{// the answer probs are evaluated in prefill mode}
			\ENDFOR
			\FOR{ $t=T$ to $1$ }
			\STATE $\sigma^\text{unnorm}_t \leftarrow R_{t+1} - R_{t} + \hat{v}_{t+1} - \hat{v}_{t}$     \textcolor{gray}{// computing the sigmas}
			\ENDFOR
			\STATE $\sigma \leftarrow \texttt{normalize} (\sigma^\text{unnorm}) $ \qquad \textcolor{gray}{// normalizing sigmas}
		\end{algorithmic}
		
	\end{algorithm}
	
	RUDDER \citep{Arjona-Medina:2019} is an algorithm for decomposing and redistributing the rewards for delayed reward problems. 
	In Deep Reinforcement Learning (DRL), it manages to mitigate the bias of the TD methods and the variance of the MCMC methods by learning an additional LSTM model \citep{Hochreiter:1997} that is then used for credit assignment to intermediate actions, speeding up the convergence of on-policy methods.
	\citet{Arjona-Medina:2019} describe the optimal reward redistribution $\Tilde{\cP}$ as such, that any future reward $\kappa(T-t-1, t) = 0$ for $0 \le t \le T - 1$.
	Under such redistribution, the following conditions are equivalent and sufficient for optimal reward redistribution:
	\begin{align}
	\kappa(T-t-1, t) &= 0 \quad \nonumber \\ \Leftrightarrow \nonumber \\ \quad
	\EXP[R_{t+1} \mid \Bs_{t-1}, a_{t-1}, \Bs_{t}, a_{t}] &= \tilde{q}^{\pi} (\Bs_{t}, a_{t}) - \tilde{q}^{\pi} (\Bs_{t-1}, a_{t-1}) \label{eq:rudder-q-form} 
	\end{align}
	We can use the action values derived for CoT MDP in Eq.~\eqref{eq:cot-mdp-action-value} together with Eq.~\eqref{eq:cot-mdp-bellman} and the second definition of optimal reward redistribution in Eq.~\eqref{eq:rudder-q-form} to derive an optimal reward redistribution for the CoT trace.
	Let us consider the scenario where the model is in state $\Bs_t$, $a_t \notin \cY$, and $a_{t-1} \notin \cY$:
	\begin{align}\label{eq:cot-mdp-redistribution}
	\EXP[R_{t+1}  & \mid \Bs_{t-1}, a_{t-1}, \Bs_{t}, a_{t}] =  \\ 
	\label{eq:cot-mdp-goes-answer-part}
	= & \; q^{\Bw} (\Bs_{t}, a_{t}) - q^{\Bw} (\Bs_{t-1}, a_{t-1}) \\
	\label{eq:cot-mdp-goes-cot-part}
	= & \; \EXP_{\By \drawnfrom p(\By \mid \Bs_{t+1})} [ \zeta(\By) ] - \EXP_{\By \drawnfrom p(\By \mid \Bs_{t})} [ \zeta(\By) ]  \\
	& + \; \EXP_{\Bu_{t+1}     } [ v^{\Bw}(\Bs_{t+2}) ] - \EXP_{u_{t}     } [ v^{\Bw}(\Bs_{t+1}) ]  \nonumber 
	\end{align}
	The part of the sum in Eq.~\eqref{eq:cot-mdp-goes-answer-part} can be computed by direct estimation of expectations in Eq.~\eqref{eq:reward_equation_cot} using the model with parameters $\Bw$. 
	To estimate these expectations under the output distribution of an autoregressive models one could use multinomial sampling as was done by \citet{Hammoud:2025}.
	At the same time, under $\{0,1\}$ outcome reward $\zeta$ and a given small set of reference answers is known, we can estimate the expectation in Eq.~\eqref{eq:reward_equation_cot} using a PR estimator.
	
	The value part of the reward redistribution (Eq.~\eqref{eq:cot-mdp-goes-cot-part}) requires estimating an expectation over the plausible subsequent steps. 
	Estimating this quantity would require sampling for every segment, which would render it either too slow to be practically applied to online training or require a considerable reduction in segment number, yielding coarser credit assignment. 
	However, this expectation would generally be low if the probabilities of the actions $p(\Bu_t \mid \Bs_t, \Bw)$ are small.
	This gives rise to our segmentation strategy in the following section, which aims to ensure that the fragments from existing CoT are as decoupled as possible.
	
	\subsection{Hybrid Segmentation Strategy}\label{sec:segment-treatment}
	
	While our algorithm can be used for token-level attribution, this would require thousands of evaluations for every trace, rendering it impractical.
	The segmentation strategy may influence the subsequent credit assignment and additional computational costs.
	Several prior works consider the segmentation of CoT traces. 
	For instance, \citet{Hammoud:2025} define subthoughts based on special sequences, e.g. ``Wait'', ``But'', etc.
	More generic substring segmentation approaches are also possible, such as splitting on double newlines.
	\citep{Guo:2025} and \citep{Gong:2026} propose segmentation strategies based on token probabilities. 
	While keyword based approaches are inherently anthropomorphic and may lead to suboptimal segmentation from the algorithmic standpoint, the purely entropy or probability based approaches can fall into the trap of synonyms, where the high immediate entropy does not lead to actual plurality of the subsequent strings. 
	
	We propose using a hybrid keyword-entropy segmentation approach (Fig.~\ref{fig:cot-mdp-and-segmentation} (Left)).
	The hybrid segmentation starts with splitting by generic keywords, such as a newline character, and later iteratively merging consecutive segments with the lowest combined entropy until the desired consolidated number of exit points is reached. 
	High entropy tokens have high plurality and are therefore opportune 'exit points' for estimating the immediate contributions. 
	Additionally, we can then homogenize segment likelihood and allow better estimation of the value term in Eq.~\eqref{eq:cot-mdp-goes-cot-part}.
	
	\subsection{Credit Assignment}
	
	Once the segmentation is done, we must assign intermediate advantages or proportions thereof to every segment.
	The goal of this stage is to construct an efficient estimator for the Eq.~\eqref{eq:cot-mdp-goes-answer-part} and \eqref{eq:cot-mdp-goes-cot-part}.
	Here, it is important to delineate credit assignment from the attribution analysis.
	The question of attribution analysis is ``what contributed to obtaining this trajectory'' whereas the question of credit assignment is "what contributed to getting this reward".
	In other words, credit assignment introduces additional information about the reward, while the attribution analysis operates merely on the sampled trajectory and environment dynamics.
	
	With the PR estimator we can estimate not only the immediate term in Eq.~\eqref{eq:cot-mdp-goes-answer-part}, but also get an estimate of the value term. 
	We can achieve that by using a reference solution path, which is usually provided in the datasets.
	Immediate reward estimates in this case can be viewed as a distance to the goal state, where the generation of the desired high utility output is inevitable. 
	
	Our PR-style estimator for the value function is then as follows:
	\begin{align}
	\hat{v}^\text{our}_{\Bw} (\Bs_{t}) 
	= & \sum_{\By, \Bu \drawnfrom q} \frac{p(\By \mid \Bu, \Bx, \Bw) p(\Bu \mid \Bx, \Bw)}{q(\By, \Bu \mid \Bx, \Bw)} \cdot \zeta(\By) \nonumber \\
	= & \frac{1}{N} \sum_{\substack{
			\By \in \{\cY^\star\} \\ 
			\Bu \in \{\cU^\star\}
	}} p(\By \mid \Bu, \Bx, \Bw) p(\Bu \mid \Bx, \Bw) \cdot \zeta(\By)     \label{eq:value-estimator-our}
	\end{align}
	Where $q$ is a proposal distribution which we define as a uniform distribution over selected reference answer-solution pairs $\cY^\star$ and $\cU^\star$.
	These sequences are a predetermined set of answers and corresponding solution paths accordingly.
	The crucial quality that the members of the set must possess is non-zero utility $\zeta$.
	In the simplest case, this set consists of a single reference answer and solution.
	Alternatively, it can feature multiple combinations of answers, solutions coming from different sources, including the model's own high utility solution answer pairs and solutions generated by teacher models.
	
	$\hat{v}^\text{our}_{\Bw}$ is an importance sampling estimator that uses the biased proposal distribution of the available solution paths.
	Let us derive the extent of bias of this point estimator using the unbiased MC estimator (for detailed transformations refer to Appx.~\ref{appx:sec:value_estimator_bias}):
	\begin{align}
	\text{Bias} & \left[ v_{\Bw}(\Bs_{t}),  \hat{v}^\text{our}_{\Bw} (\Bs_{t})\right] = \EXP[\hat{v}^\text{our}_{\Bw} (\Bs_{t})] - v_{\Bw}(\Bs_{t}) \nonumber \\
	= & - \sum_{\By \in \cY \setminus \cY^\star, \Bu \in \cU \setminus \cU^\star} \frac{1}{N} \underbrace{p(\By, \Bu \mid \Bx, \Bw)}_{\text{ answer \& solution prob.}} \cdot \underbrace{\zeta(\By)}_{\text{utility}}  \label{eq:value_estimator_bias}
	\end{align}
	where $Z$ is the cardinality of the set of all possible solutions and answers for the autoregressive model and $N$ is the cardinality of the reference solution set. 
	We note that $Z$, being the number of all possible sequences generated by the model, is finite since the maximum length must be specified for the autoregressive decoding algorithm \citep{Malinin:2020}.
	Essentially, this means that for every pair $\By^\star$ and $\Bu^\star$ we sum the integrand over all continuations that are not this specific one.
	Under the practically sensible assumption that $\zeta$ is non-negative, this bias is less or equal to zero and will lead to underestimation of the value function. 
	In the best case scenario, our proposal distribution $q$ captures all sequences that have non-zero probability and non-zero utility leading to zero bias. 
	
	While explicitly estimating the bias in Eq.~\eqref{eq:value_estimator_bias} is intractable, we can draw some hypotheses about it by leveraging the empirical insights from Bayesian methods for LLMs. 
	It is known that the predictive distribution of Language Models is structured and will generally contain clusters that are correlated \citep{Kuhn:2023,Farquhar:2024}.
	With this in mind, if we take a difference between $\hat{v}^\text{our}_{\Bw}(\cdot)$ evaluated at $\Bs_{t}$ and $\Bs_{t+1}$ as is required to estimate the Eq.~\eqref{eq:cot-mdp-goes-cot-part}, we can expect the difference of the probabilities of the reference solution to be indicative of the dynamics for the whole cluster. 
	One scenario where this bias could be substantial is when there are multiple diverse solutions that are plausible under the model $\Bw$ and have high utility for the given problem $\Bx$.
	If the solutions are diverse, they might be parts of different clusters that are disjoint in terms of probability.
	In this case, the mass in Eq.~\eqref{eq:value_estimator_bias} would have a high magnitude and it would be possible that the estimates obtained through the difference of $\hat{v}^\text{our}_{\Bw}(\cdot)$ could even be nonsensical.
	
	\paragraph{Subgoal structure of reference solution.}
	We assume that the regions of the predictive distribution of the reasoning model that are disjoint from the region containing the reference solution, yet contain sequences with positive utility are rare.
	Given the body of work in Bayesian Language Modeling (i.e., \citet{Aichberger:2024a}) we feel such an assumption is safe.
	\citep{DeepSeek-AI:2025} and the majority of subsequent RLVR works consider mathematics datasets (e.g. \citet{aime24,Hendrycks:2021a}) which have the advantage of having unique solutions (down to symbolic transformation).
	Non-mathematical problems, such as logic and programming \citep{Stojanovski:2025, Hendrycks:2021}, also provide unique solutions down to invariances.
	In many problems with composite answers (i.e., answers that contain multiple parts that can be independently correct or wrong), such as Sudoku puzzles or Python programs, the subgoal structure is revealed even without an explicit solution path.
	For example, in a Sudoku puzzle, the placement of each number is a subtask and when we see the filled-out answer to it, we can assess the achievement of individual subgoals.
	
	Often a reference solution path is not available. 
	As we show in the later section, the reward redistribution requires more information than just the produced sequence itself.
	The manner in which such information is obtained (whether the reference solution or an extensive tree search) is beyond the scope of our work.
	In this case one can consider starting with applying a search algorithm of choice which attempts to find a solution for a given problem.
	This solution path can then be summarized and used for redistribution.
	We assume that some hint of the subgoal structure is available during training.
	
	\subsection{Integrating Reward Redistribution into Commonly Used RL Objectives}
	
	\begin{table*}    \centering
		\setlength{\tabcolsep}{5pt}
		\caption{\methodname{} performance improvements on Numina-CoT \citep{Numina:2024} dataset with long generation length ($25$k tokens). \methodname{} yields greater improvement than GRPO.}
		\label{tab:res:online_medium_scale_results}
		\begin{tabular}{lccccc}
			\toprule
			Model & AIME24 & AIME25 & AIME26 & Minerva & MATH500 \\ 
			\toprule
			Qwen3-4B Instruct 2507 (starting point) & $.692$ & $.429$ & $.267$ & $.906$ & $.781$ \\
			GRPO &  $.850$ & $\mathbf{.600}$ & $.442$ & $.915$ & $.804$ \\
			\methodname{} &  $\mathbf{.908}$ & $.583$ & $\mathbf{.475}$ & $\mathbf{.935}$ & $\mathbf{.823}$ \\
			\bottomrule \\
		\end{tabular}
	\end{table*}
	
	\begin{table*}[b]
		\centering
		\setlength{\tabcolsep}{5pt}
		\caption{\methodname{} models, small-scale application of the reward redistribution to online refinement of small reasoning models. Notably, the \methodname{} models were tuned with the context size of only $1024$. The tuning was performed on the \texttt{open-rs} dataset \citep{Dang:2025}, while the \texttt{open-rs} models were taken as is from HuggingFace.}
		\label{tab:res:online_small_scale_results}
		\begin{tabular}{lccccc}
			\toprule
			Model & AIME24 & AMC23 & MATH500 & MINERVA & \begin{tabular}{@{}c@{}}Olympiad \\ Bench\end{tabular} \\ 
			\toprule
			Qwen2.5-1.5B R1-Distilled (base) & $.300$ & $.675$ & $.840$ & $\mathbf{.320}$ & $.521$\\
			Open-RS (3 checkpoints) & $\mathbf{.366}$ & $\underline{.725}$ & $\underline{.848}$ & $.316$ & $\underline{.524}$ \\
			\methodname{}-Nano (2 checkpoints) & $\underline{.333}$ & $\mathbf{.800}$ & $\mathbf{.858}$ & $.305$ & $\mathbf{.531}$ \\
			\bottomrule \\
		\end{tabular}
	\end{table*}
	
	The \methodname{} redistribution approximation derived so far depends only on the properties of the CoT generation MDP.
	This means that it can be integrated into any RL objective so long as it is applied to a problem of that MDP structure. 
	One important condition of reward redistribution is return-equivalence, meaning that the original episode return must be preserved in the reward redistribution SDP. 
	We further reformulate the generalized policy gradient objective from \citet{Shao:2024} to include the redistribution of rewards:
	\begin{align} 
	\nabla_\theta & \cJ_{\cA}(\theta) = \EXP \underbrace{_{(q,o) \drawnfrom \cD}}_{\text{Data Source}} \bigg[ \sum_{t=0}^{|o|} \underbrace{ \sigma(t, q, o, \pi_\text{rf})}_{\text{Redistribution}} \underbrace{GC_{\cA} (q,o,\pi_\text{rf})}_{\text{Gradient Coefficient}} \underbrace{\nabla_\theta \log \pi_\theta (o_t \mid q, o_{<t})}_{\text{Token-Wise Gradient}} \label{eq:deepseek-math-objective-with-redistribution}
	\bigg]
	\end{align} 
	Where $\sigma (t, q, o, \pi_\text{rf})$ is the token-wise reward coefficient that depends on the reference model and question-answer pair.
	The policy gradient formulations attribute reward uniformly over the whole sequence Eq.~\eqref{eq:deepseek-math-objective-uniform-sigma}:
	\begin{align}
	\label{eq:deepseek-math-objective-uniform-sigma}
	\sigma(t, q, o, \pi_\text{rf}) = \frac{1}{|o|}
	\end{align}
	$\sigma$ must sum up to $1$ in order to satisfy the return equivalence from \citet{Arjona-Medina:2019}.
	We note, that this formulation does not bias the original objective according to Theorem 1 in \citet{Arjona-Medina:2019} if the condition holds.
	This leads to the same optimal policy.
	In practice, the diversity of data and breadth of autoregressive predictive distribution mean that full convergence i.e. perfect knowledge of the general reasoning, is unattainable in practice.
	As a consequence, speeding up the convergence rate would mean de facto improvement of final performance.
	
	Several GRPO improvements use the normalizer $\sigma$ for reward shaping without adhering to these conditions, therefore changing the optimal policy.
	For example, BNPO \citep{Xiao:2025} uses $\sigma = \frac{1}{|\cB|}$ where $\cB$ is the number of tokens in a batch.
	DR-GRPO \citep{Liu:2025} uses $\sigma = \frac{1}{|\cM|}$ where $\cM$ is the maximum number of tokens allowed in the batch.
	
	\section{Experiments}
	
	In this section, we investigate some of the empirical properties relating to the proposed \methodname{}.
	The implementation details of the experiments in this section are provided in Appx.~\ref{appx:sec:experiment_details}.
	
	\paragraph{Variance and Bias of Truncated MC Value Estimator.}
	
	\begin{figure}
		\centering
		\includegraphics[width=.9\linewidth]{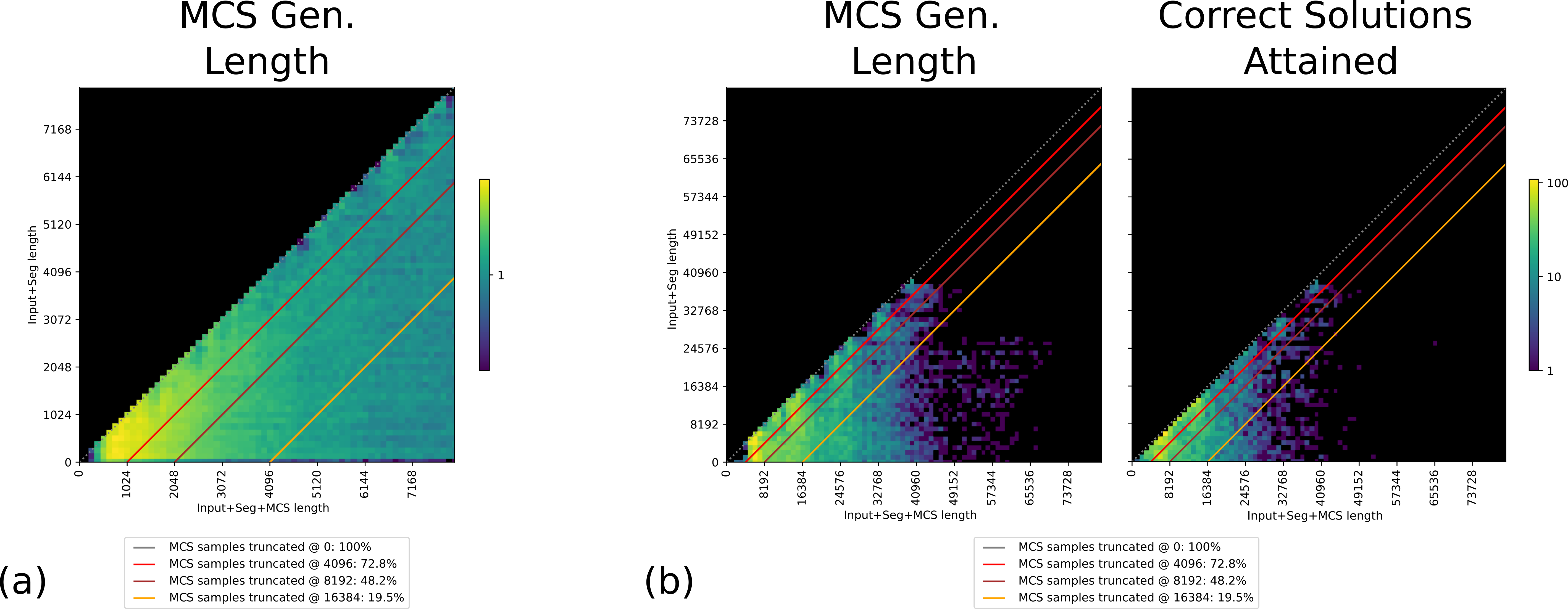}
		\caption{
			Proportion of rollouts lost by the truncated MC sampling with respect to truncation length.
			(a) MATH-500 dataset; (b) AIME-25 dataset. 
			Computing MC sampling estimates in (b) took $100$ GPU-hours for the $30$ questions given and maximum number of segments of $40$.
			In both cases we see that truncation of the MC generations leads to loss of substantial amount of the completions, including those that lead to correct answers.
		}
		\label{fig:mcts_truncation_bias}
	\end{figure}
	
	We conducted an analysis of the variability of the estimates of intermediate values by MC sampling.
	In Appx.Fig.~\ref{fig:mcts_sd_bootstrap}, we show how the number of MC samples impacts the standard deviation of the estimates.
	We observe SD of the estimator falling below the $0.1$ mark after approximately $5$ samples.
	In this example, we have computed the MC values by generating sequences with a fixed total horizon, so the effects of increased variance at the early stages of CoT generation are not visible. 
	Such computation would be impractical at these levels of granularity for on-policy optimization.
	In Fig.~\ref{fig:mcts_truncation_bias} we show that truncating the MC sampling completions can lead to loss of a large number of completions, including ones that lead to correct answers.
	This introduces a bias similar to that in Eq.~\eqref{eq:value_estimator_bias}.
	
	\paragraph{Correlation analysis between attribution and credit assignment methods.}
	
\begin{figure}[t]
		\centering
		\includegraphics[width=0.6\linewidth]{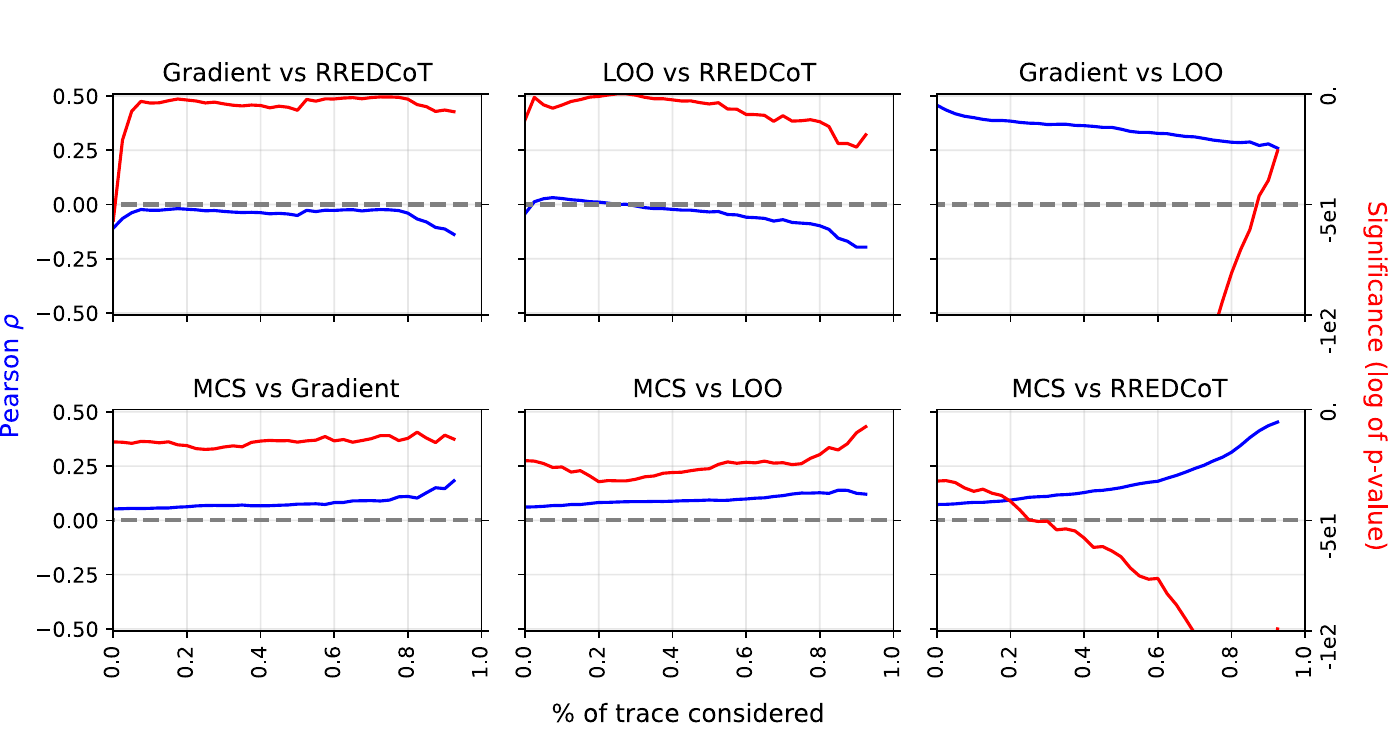}
		\caption{
			Correlation between LOO, Gradient attribution, MC sampling and RREDCoT credit assignment techniques.
			On the x-axis we have the percent of the trace used from the right, e.g., a value of $0.25$ means that from each trace, the first 25\    Blue lines are correlation coefficients (left scale) while the red lines are their corresponding log(p-values) (right scale). 
			All values were computed using the model that produced the CoT.
			The gradient norm and LOO, being attribution methods, correlate highly with each other, while RREDCoT, being a credit assignment method, is most similar to MC sampling, especially at the later stages. 
		}
		\label{fig:plot_loo_grad_rredcot_mcts}
	\end{figure}

	A number of attribution techniques exist, such as Shapley values \citep{Beechey:2023}, Leave One Out (LOO) \citep{Liu:2025e, Khandoga:2026} and gradient-based techniques \citep{Sundararajan:2017}.
	Attribution methods have their advantages and drawbacks, and there is no one-size-fits-all.
	Generally, attribution in neural networks is a computationally intensive procedure.
	We want to test the hypothesis that attribution unconditioned on additional information is disjoint from credit assignment/reward redistribution and would be ill-suited for explicitly weighing the tokens during training.
	The results of the correlation analysis are presented in Fig.~\ref{fig:plot_loo_grad_rredcot_mcts}.
	The gradient attribution is highly correlated with LOO attribution, meaning that using the LOO attribution values as an explicit weighing signal at training time would be redundant.
	This shows that using LOO style attribution analysis for the purposes of assigning intermediate rewards would be redundant at training time, since much of its allocation would be implicitly performed by the gradient descent.
	We speculate that this would also negate the effect of any other purely attribution explicit weight assignment, including, for example, attention weight-based attribution. 
	At the same time, RREDCoT reward redistribution shows relatively high correlation, especially towards the later parts of the trajectories, where the MC estimates might be more precise.
	
	\paragraph{On Policy LM Fine-Tuning.}
	
	We have utilized the objective in Eq.~\eqref{eq:deepseek-math-objective-with-redistribution} to assess the utility of our reward redistribution.
	We used 4B parameter Qwen3 Instruct model as a starting point with maximum generation length set to $25$k tokens.
	The optimization was performed for $500$ steps with hyperparameters detailed in Appx.\ref{appx:sec:experiment_details}.
	The $\sigma$ values for redistribution were normalized using L1 norm. 
	This aligns conveniently with using PR approaches \citep{Yu:2025a, Zhou:2025} to estimate the reward, as the sum of all intermediate rewards in log space equals the log PR.
	For GRPO, standard verifiers along were used along with several beneficial adjustments, such as sequence level importance \citep{Zheng:2025} and implementation level importance weights as is provided in the Transformer Reinforcement Learning library \citep{vonWerra:2020}.
	
	For each question in evaluation datasets $8$ independent rollouts have been produced.
	Each output was evaluated using an ensemble of the LLM-as-a-Judge models with two prompts $8$ samples each \citep{Ielanskyi:2026} for robustness. 
	Tab.~\ref{tab:res:online_medium_scale_results} shows a confident increase in efficiency when using our method for longer generation lengths.
	
	\paragraph{Redistributing Reward Using Model's Own Answer.}\label{sec:bootstrapping_with_own_answers}
	
	In order to check if our objective works with bootstrapping the model on its own answers, we have utilized the objective in Eq.~\eqref{eq:deepseek-math-objective-with-redistribution} but used the models own answer for redistribution.
	The advantage terms were estimated as in GRPO (Eq.~\eqref{appx:eq:ry_with_verifiers_mc_estimate}).
	Since the advantage values were computed from the models own answers, we could normalize the $\sigma$ values using the softmax, making the $\sigma$ values positive and sum up to $1$.
	Log-differences of transitions were used as attribution values to estimate the optimal redistribution (Eq.~\eqref{eq:cot-mdp-goes-answer-part}).
	
	The results are presented in Tab.~\ref{tab:res:online_small_scale_results}.
	Best of $2$ checkpoints were taken for RREDCoT and best of $3$ for open-rs over a short training run (300 steps). 
	The resulting model compares favorably to those obtained by \citep{Dang:2025} using GRPO and hyperparameter tuning. 
	
	\section{Limitations}
	Our proposed method relies on the knowledge of the solution to the problem and some degree of subgoal structure.
	Normally this would be an existing solution trace that is not atypical under the distribution of general text. 
	While a vast majority of the post-training datasets known to us provide such information, it cannot be taken for granted.
	Our method cannot be simply applied to the problems for which no hint of solution path is available or the provided reference solution is uninformative.
	One category of such problems is constraint satisfaction problems.
	Mitigation of this could be attempted with the use of bootstrapping on the models own correct answers (as was assessed in the last paragraph (Sec.4/P.\ref{sec:bootstrapping_with_own_answers}) of the preceding section). 
	This, however, will only work if a positive signal is attainable by the model and precludes the use of PR based approaches to sequence level advantage estimation.
	
	Furthermore our method involves additional computational overhead compared to the simpler GRPO variants.
	In practice our optimization runs with \methodname{} required $1.5-2$x the amount of compute as the runs without redistribution.
	This is still much faster compared to the MC sampling based intermediate value estimates and we believe that it is a reasonable trade-off. 
	Our implementation of value function estimation for the intermediate values was aggressively optimized for reuse of KV values and batched inference. 
	The peak GPU memory utilization was unchanged from naive GRPO.
	
	\section{Conclusion and Outlook}\label{sec:discussion}
	
	In this work, we introduce a new reward redistribution and credit assignment method \methodname{} which is intended to improve the sample efficiency of reasoning models fine-tuning. 
	We have described the method and its versatile compatibility with the existing RLVR / RLPR pipelines and optimization objectives, conducted evaluations regarding its empirical benefits and discussed its relation to alternatives.
	The estimator agrees well with the values provided by our thorough MC sampling estimate while being considerably less computationally expensive.
	We additionally show that traditional attribution methods capture different aspects of CoT generation than reward redistribution and MC-based credit assignment.
	These insights show that attribution methods that are not conditioned on the correct solutions are bound to capture aspects not related to arriving at the desired state.
	We anticipate future work to further explore the application of Bayesian methods to credit assignment problems in this increasingly important setting.
	The \methodname{} can function as both a training signal amplifier and an analysis tool.
	
	\clearpage
	
	\section*{Acknowledgments}
	
	The ELLIS Unit Linz, the LIT AI Lab, the Institute for Machine Learning, are supported by the Federal State Upper Austria. We thank the projects FWF AIRI FG 9-N (10.55776/FG9), AI4GreenHeatingGrids (FFG- 899943), Stars4Waters (HORIZON-CL6-2021-CLIMATE-01-01), FWF Bilateral Artificial Intelligence (10.55776/COE12). We thank NXAI GmbH, Audi AG, Silicon Austria Labs (SAL), Merck Healthcare KGaA, GLS (Univ. Waterloo), T\"{U}V Holding GmbH, Software Competence Center Hagenberg GmbH, dSPACE GmbH, TRUMPF SE + Co. KG.
	
	\bibliography{arxiv_refs}

@inproceedings{Arjona-Medina:2019,
  title = {{{RUDDER}}: {{Return}} Decomposition for Delayed Rewards},
  booktitle = {Advances in Neural Information Processing Systems},
  author = {Arjona-Medina, Jose A. and Gillhofer, Michael and Widrich, Michael and Unterthiner, Thomas and Brandstetter, Johannes and Hochreiter, Sepp},
  year = {2019},
  volume = {32},
  url = {https://proceedings.neurips.cc/paper_files/paper/2019/file/16105fb9cc614fc29e1bda00dab60d41-Paper.pdf}
}

@misc{Chandak:2025,
  title = {Incorrect {{Baseline Evaluations Call}} into {{Question Recent LLM-RL Claims}} | {{Notion}}},
  author = {Chandak, Nikhil and Goel, Shashwat and Ameya, Prabhu},
  year = {2025},
  url = {https://safe-lip-9a8.notion.site/Incorrect-Baseline-Evaluations-Call-into-Question-Recent-LLM-RL-Claims-2012f1fbf0ee8094ab8ded1953c15a37},
  organization = {safe-lip-9a8 on Notion}
}

@misc{Dang:2025,
  title = {Reinforcement {{Learning}} for {{Reasoning}} in {{Small LLMs}}: {{What Works}} and {{What Doesn}}'t},
  shorttitle = {Reinforcement {{Learning}} for {{Reasoning}} in {{Small LLMs}}},
  author = {Dang, Quy-Anh and Ngo, Chris},
  date = {2025-03-20},
  year = {2025},
  eprint = {2503.16219},
  eprinttype = {arXiv},
  eprintclass = {cs},
  url = {http://arxiv.org/abs/2503.16219},
  pubstate = {prepublished}
}

@misc{DeepSeek-AI:2025,
  title = {{{DeepSeek-R1}}: {{Incentivizing Reasoning Capability}} in {{LLMs}} via {{Reinforcement Learning}}},
  shorttitle = {{{DeepSeek-R1}}},
  author = {DeepSeek-AI and Guo, Daya and Yang, Dejian and Zhang, Haowei and Song, Junxiao and Zhang, Ruoyu and Xu, Runxin and Zhu, Qihao and Ma, Shirong and Wang, Peiyi and Bi, Xiao and Zhang, Xiaokang and Yu, Xingkai and Wu, Yu and Wu, Z. F. and Gou, Zhibin and Shao, Zhihong and {et al.}},
  date = {2025-01-22},
  year = {2025},
  eprint = {2501.12948},
  eprinttype = {arXiv},
  eprintclass = {cs},
  url = {http://arxiv.org/abs/2501.12948},
  pubstate = {prepublished}
}

@misc{Gong:2026,
  title = {Segmental {{Advantage Estimation}}: {{Enhancing PPO}} for {{Long-Context LLM Training}}},
  shorttitle = {Segmental {{Advantage Estimation}}},
  author = {Gong, Xue and Yi, Qi and Nan, Ziyuan and Huang, Guanhua and Li, Kejiao and Jiang, Yuhao and Xiong, Ruibin and Xu, Zenan and Guo, Jiaming and Peng, Shaohui and Zhou, Bo},
  date = {2026-01-12},
  year = {2026},
  eprint = {2601.07320},
  eprinttype = {arXiv},
  eprintclass = {cs},
  url = {http://arxiv.org/abs/2601.07320},
  pubstate = {prepublished}
}

@misc{Guo:2025,
  title = {Segment {{Policy Optimization}}: {{Effective Segment-Level Credit Assignment}} in {{RL}} for {{Large Language Models}}},
  shorttitle = {Segment {{Policy Optimization}}},
  author = {Guo, Yiran and Xu, Lijie and Liu, Jie and Ye, Dan and Qiu, Shuang},
  date = {2025-10-21},
  year = {2025},
  eprint = {2505.23564},
  eprinttype = {arXiv},
  eprintclass = {cs},
  url = {http://arxiv.org/abs/2505.23564},
  pubstate = {prepublished}
}

@misc{Hammoud:2025,
  title = {Beyond the {{Last Answer}}: {{Your Reasoning Trace Uncovers More}} than {{You Think}}},
  shorttitle = {Beyond the {{Last Answer}}},
  author = {Hammoud, Hasan Abed Al Kader and Itani, Hani and Ghanem, Bernard},
  date = {2025-04-29},
  year = {2025},
  eprint = {2504.20708},
  eprinttype = {arXiv},
  eprintclass = {cs},
  url = {http://arxiv.org/abs/2504.20708},
  pubstate = {prepublished}
}

@misc{Jayalath:2025,
  title = {Compute as {{Teacher}}: {{Turning Inference Compute Into Reference-Free Supervision}}},
  shorttitle = {Compute as {{Teacher}}},
  author = {Jayalath, Dulhan and Goel, Shashwat and Foster, Thomas and Jain, Parag and Gururangan, Suchin and Zhang, Cheng and Goyal, Anirudh and Schelten, Alan},
  date = {2025-09-17},
  year = {2025},
  eprint = {2509.14234},
  eprinttype = {arXiv},
  eprintclass = {cs},
  url = {http://arxiv.org/abs/2509.14234},
  pubstate = {prepublished},
  version = {1}
}

@misc{Kazemnejad:2025,
  title = {{{VinePPO}}: {{Refining Credit Assignment}} in {{RL Training}} of {{LLMs}}},
  shorttitle = {{{VinePPO}}},
  author = {Kazemnejad, Amirhossein and Aghajohari, Milad and Portelance, Eva and Sordoni, Alessandro and Reddy, Siva and Courville, Aaron and Roux, Nicolas Le},
  date = {2025-06-03},
  year = {2025},
  eprint = {2410.01679},
  eprinttype = {arXiv},
  eprintclass = {cs},
  url = {http://arxiv.org/abs/2410.01679},
  pubstate = {prepublished}
}

@inproceedings{Lewkowycz:2022,
  title = {Solving {{Quantitative Reasoning Problems}} with {{Language Models}}},
  author = {Lewkowycz, Aitor and Andreassen, Anders Johan and Dohan, David and Dyer, Ethan and Michalewski, Henryk and Ramasesh, Vinay Venkatesh and Slone, Ambrose and Anil, Cem and Schlag, Imanol and Gutman-Solo, Theo and Wu, Yuhuai and Neyshabur, Behnam and Gur-Ari, Guy and Misra, Vedant},
  date = {2022-10-31},
  year = {2022},
  url = {https://openreview.net/forum?id=IFXTZERXdM7},
  booktitle = {Advances in {{Neural Information Processing Systems}}}
}

@misc{Li:2026,
  title = {Outcome-{{Grounded Advantage Reshaping}} for {{Fine-Grained Credit Assignment}} in {{Mathematical Reasoning}}},
  author = {Li, Ziheng and Kang, Liu and Xiao, Feng and Xing, Luxi and Si, Qingyi and Li, Zhuoran and Gong, Weikang and Yang, Deqing and Xiao, Yanghua and Guo, Hongcheng},
  date = {2026-01-12},
  year = {2026},
  eprint = {2601.07408},
  eprinttype = {arXiv},
  eprintclass = {cs},
  url = {http://arxiv.org/abs/2601.07408},
  pubstate = {prepublished}
}

@misc{Liu:2025,
  title = {Understanding {{R1-Zero-Like Training}}: {{A Critical Perspective}}},
  shorttitle = {Understanding {{R1-Zero-Like Training}}},
  author = {Liu, Zichen and Chen, Changyu and Li, Wenjun and Qi, Penghui and Pang, Tianyu and Du, Chao and Lee, Wee Sun and Lin, Min},
  date = {2025-03-26},
  year = {2025},
  eprint = {2503.20783},
  eprinttype = {arXiv},
  eprintclass = {cs},
  url = {http://arxiv.org/abs/2503.20783},
  pubstate = {prepublished}
}

@misc{Muennighoff:2025,
  title = {S1: {{Simple}} Test-Time Scaling},
  shorttitle = {S1},
  author = {Muennighoff, Niklas and Yang, Zitong and Shi, Weijia and Li, Xiang Lisa and Fei-Fei, Li and Hajishirzi, Hannaneh and Zettlemoyer, Luke and Liang, Percy and Candès, Emmanuel and Hashimoto, Tatsunori},
  date = {2025-01-31},
  year = {2025},
  eprint = {2501.19393},
  eprinttype = {arXiv},
  eprintclass = {cs},
  url = {http://arxiv.org/abs/2501.19393},
  pubstate = {prepublished}
}

@misc{Ou:2025,
  title = {{{SERL}}: {{Self-Examining Reinforcement Learning}} on {{Open-Domain}}},
  shorttitle = {{{SERL}}},
  author = {Ou, Weixuan and Zheng, Yanzhao and Sun, Shuoshuo and Zhang, Wei and Dong, Baohua and Zhu, Hangcheng and Huang, Ruohui and Yu, Gang and Yan, Pengwei and Qiao, Yifan},
  date = {2025-11-18},
  year = {2025},
  eprint = {2511.07922},
  eprinttype = {arXiv},
  eprintclass = {cs},
  url = {http://arxiv.org/abs/2511.07922},
  pubstate = {prepublished}
}

@misc{Shao:2024,
  title = {{{DeepSeekMath}}: {{Pushing}} the {{Limits}} of {{Mathematical Reasoning}} in {{Open Language Models}}},
  shorttitle = {{{DeepSeekMath}}},
  author = {Shao, Zhihong and Wang, Peiyi and Zhu, Qihao and Xu, Runxin and Song, Junxiao and Bi, Xiao and Zhang, Haowei and Zhang, Mingchuan and Li, Y. K. and Wu, Y. and Guo, Daya},
  date = {2024-04-27},
  year = {2024},
  eprint = {2402.03300},
  eprinttype = {arXiv},
  eprintclass = {cs},
  url = {http://arxiv.org/abs/2402.03300},
  pubstate = {prepublished}
}

@misc{Xiao:2025,
  title = {{{BNPO}}: {{Beta Normalization Policy Optimization}}},
  shorttitle = {{{BNPO}}},
  author = {Xiao, Changyi and Zhang, Mengdi and Cao, Yixin},
  date = {2025-06-03},
  year = {2025},
  eprint = {2506.02864},
  eprinttype = {arXiv},
  eprintclass = {cs},
  url = {http://arxiv.org/abs/2506.02864},
  pubstate = {prepublished}
}

@misc{Xie:2025a,
  title = {{{CAPO}}: {{Towards Enhancing LLM Reasoning}} through {{Verifiable Generative Credit Assignment}}},
  shorttitle = {{{CAPO}}},
  author = {Xie, Guofu and Shi, Yunsheng and Tian, Hongtao and Yao, Ting and Zhang, Xiao},
  date = {2025-08-04},
  year = {2025},
  eprint = {2508.02298},
  eprinttype = {arXiv},
  eprintclass = {cs},
  url = {http://arxiv.org/abs/2508.02298},
  pubstate = {prepublished}
}

@misc{Yang:2025,
  title = {Qwen3 {{Technical Report}}},
  author = {Yang, An and Li, Anfeng and Yang, Baosong and Zhang, Beichen and Hui, Binyuan and Zheng, Bo and Yu, Bowen and Gao, Chang and Huang, Chengen and Lv, Chenxu and Zheng, Chujie and Liu, Dayiheng and {et al.}},
  date = {2025-05-14},
  year = {2025},
  eprint = {2505.09388},
  eprinttype = {arXiv},
  eprintclass = {cs},
  url = {http://arxiv.org/abs/2505.09388},
  pubstate = {prepublished}
}

@misc{Yu:2025,
  title = {{{DAPO}}: {{An Open-Source LLM Reinforcement Learning System}} at {{Scale}}},
  shorttitle = {{{DAPO}}},
  author = {Yu, Qiying and Zhang, Zheng and Zhu, Ruofei and Yuan, Yufeng and Zuo, Xiaochen and Yue, Yu and Fan, Tiantian and Liu, Gaohong and Liu, Lingjun and Liu, Xin and Lin, Haibin and Lin, Zhiqi and Ma, Bole and Sheng, Guangming and Tong, Yuxuan and Zhang, Chi and {et al.}},
  date = {2025-03-18},
  year = {2025},
  eprint = {2503.14476},
  eprinttype = {arXiv},
  eprintclass = {cs},
  url = {http://arxiv.org/abs/2503.14476},
  pubstate = {prepublished}
}

@misc{Yu:2025a,
  title = {{{RLPR}}: {{Extrapolating RLVR}} to {{General Domains}} without {{Verifiers}}},
  shorttitle = {{{RLPR}}},
  author = {Yu, Tianyu and Ji, Bo and Wang, Shouli and Yao, Shu and Wang, Zefan and Cui, Ganqu and Yuan, Lifan and Ding, Ning and Yao, Yuan and Liu, Zhiyuan and Sun, Maosong and Chua, Tat-Seng},
  date = {2025-06-23},
  year = {2025},
  eprint = {2506.18254},
  eprinttype = {arXiv},
  eprintclass = {cs},
  url = {http://arxiv.org/abs/2506.18254},
  pubstate = {prepublished}
}

@misc{Zheng:2025,
  title = {Group {{Sequence Policy Optimization}}},
  author = {Zheng, Chujie and Liu, Shixuan and Li, Mingze and Chen, Xiong-Hui and Yu, Bowen and Gao, Chang and Dang, Kai and Liu, Yuqiong and Men, Rui and Yang, An and Zhou, Jingren and Lin, Junyang},
  date = {2025-07-28},
  year = {2025},
  eprint = {2507.18071},
  eprinttype = {arXiv},
  eprintclass = {cs},
  url = {http://arxiv.org/abs/2507.18071},
  pubstate = {prepublished}
}

@misc{Zhou:2025,
  title = {Reinforcing {{General Reasoning}} without {{Verifiers}}},
  author = {Zhou, Xiangxin and Liu, Zichen and Sims, Anya and Wang, Haonan and Pang, Tianyu and Li, Chongxuan and Wang, Liang and Lin, Min and Du, Chao},
  date = {2025-05-27},
  year = {2025},
  eprint = {2505.21493},
  eprinttype = {arXiv},
  eprintclass = {cs},
  url = {http://arxiv.org/abs/2505.21493},
  pubstate = {prepublished}
}

@article{Hochreiter:1997,
  title = {Long {{Short-Term Memory}}},
  author = {Hochreiter, Sepp and Schmidhuber, Jürgen},
  year = {1997},
  journal = {Neural Computation},
  volume = {9},
  pages = {1735--1780}
}

@inproceedings{Sundararajan:2017,
  title = {Axiomatic {{Attribution}} for {{Deep Networks}}},
  booktitle = {Proceedings of the 34th {{International Conference}} on {{Machine Learning}}},
  author = {Sundararajan, Mukund and Taly, Ankur and Yan, Qiqi},
  date = {2017-07-17},
  year = {2017},
  pages = {3319--3328},
  url = {https://proceedings.mlr.press/v70/sundararajan17a.html},
  eventtitle = {International {{Conference}} on {{Machine Learning}}}
}

@misc{Aichberger:2024a,
  title = {Rethinking {{Uncertainty Estimation}} in {{Natural Language Generation}}},
  author = {Aichberger, Lukas and Schweighofer, Kajetan and Hochreiter, Sepp},
  date = {2024-12-19},
  year = {2024},
  eprint = {2412.15176},
  eprinttype = {arXiv},
  eprintclass = {cs},
  url = {http://arxiv.org/abs/2412.15176},
  pubstate = {prepublished}
}

@misc{aime24,
      title={American Invitational Mathematics Examination (AIME)}, 
      author={Zhang, Yifan and Math-AI, Team},
      year={2024},
}

@article{Hendrycks:2021a,
  title = {Measuring {{Mathematical Problem Solving With}} the {{MATH Dataset}}},
  author = {Hendrycks, Dan and Burns, Collin and Kadavath, Saurav and Arora, Akul and Basart, Steven and Tang, Eric and Song, Dawn and Steinhardt, Jacob},
  year = {2021},
  journal = {Proceedings of the Neural Information Processing Systems Track on Datasets and Benchmarks},
  volume = {1}
}

@article{Hendrycks:2021,
  title = {Measuring {{Coding Challenge Competence With APPS}}},
  author = {Hendrycks, Dan and Basart, Steven and Kadavath, Saurav and Mazeika, Mantas and Arora, Akul and Guo, Ethan and Burns, Collin and Puranik, Samir and He, Horace and Song, Dawn and Steinhardt, Jacob},
  year = {2021},
  journal = {Proceedings of the Neural Information Processing Systems Track on Datasets and Benchmarks},
  volume = {1}
}

@misc{Stojanovski:2025,
  title = {{{REASONING GYM}}: {{Reasoning Environments}} for {{Reinforcement Learning}} with {{Verifiable Rewards}}},
  shorttitle = {{{REASONING GYM}}},
  author = {Stojanovski, Zafir and Stanley, Oliver and Sharratt, Joe and Jones, Richard and Adefioye, Abdulhakeem and Kaddour, Jean and K{\"o}pf, Andreas},
  year = {2025},
  number = {arXiv:2505.24760},
  eprint = {2505.24760},
  url = {https://arxiv.org/abs/2505.24760}
}

@inproceedings{Ouyang:2022,
  title = {Training Language Models to Follow Instructions with Human Feedback},
  author = {Ouyang, Long and Wu, Jeffrey and Jiang, Xu and Almeida, Diogo and Wainwright, Carroll and Mishkin, Pamela and Zhang, Chong and Agarwal, Sandhini and Slama, Katarina and Ray, Alex and Schulman, John and Hilton, Jacob and Kelton, Fraser and Miller, Luke and Simens, Maddie and Askell, Amanda and Welinder, Peter and Christiano, Paul F. and Leike, Jan and Lowe, Ryan},
  date = {2022-12-06},
  year = {2022},
  booktitle = {Advances in Neural Information Processing Systems},
  volume = {35},
  pages = {27730--27744},
  url = {https://proceedings.neurips.cc/paper_files/paper/2022/hash/b1efde53be364a73914f58805a001731-Abstract-Conference.html}
}

@misc{Team:2025,
  title = {Kimi K1.5: {{Scaling Reinforcement Learning}} with {{LLMs}}},
  shorttitle = {Kimi K1.5},
  author = {{Kimi Team} and Du, Angang and Gao, Bofei and Xing, Bowei and Jiang, Changjiu and Chen, Cheng and Li, Cheng and Xiao, Chenjun and Du, Chenzhuang and Liao, Chonghua and Tang, Chuning and Wang, Congcong and Zhang, Dehao and Yuan, Enming and {et al.}},
  year = {2025},
  number = {arXiv:2501.12599},
  eprint = {2501.12599},
  url = {https://arxiv.org/abs/2501.12599}
}

@misc{Schulman:17,
  title = {Proximal {{Policy Optimization Algorithms}}},
  author = {Schulman, John and Wolski, Filip and Dhariwal, Prafulla and Radford, Alec and Klimov, Oleg},
  date = {2017-08-28},
  year = {2017},
  eprint = {1707.06347},
  eprinttype = {arXiv},
  eprintclass = {cs},
  url = {http://arxiv.org/abs/1707.06347},
  pubstate = {prepublished}
}

@misc{Shao:2025,
  title = {Spurious {{Rewards}}: {{Rethinking Training Signals}} in {{RLVR}}},
  shorttitle = {Spurious {{Rewards}}},
  author = {Shao, Rulin and Li, Shuyue Stella and Xin, Rui and Geng, Scott and Wang, Yiping and Oh, Sewoong and Du, Simon Shaolei and Lambert, Nathan and Min, Sewon and Krishna, Ranjay and Tsvetkov, Yulia and Hajishirzi, Hannaneh and Koh, Pang Wei and Zettlemoyer, Luke},
  year = {2025},
  number = {arXiv:2506.10947},
  eprint = {2506.10947},
  url = {https://arxiv.org/abs/2506.10947}
}

@inproceedings{Malinin:2020,
  title = {Uncertainty {{Estimation}} in {{Autoregressive Structured Prediction}}},
  author = {Malinin, Andrey and Gales, Mark},
  date = {2020-10-02},
  year = {2020},
  url = {https://openreview.net/forum?id=jN5y-zb5Q7m},
  booktitle = {International {{Conference}} on {{Learning Representations}}}
}

@misc{Liu:2025e,
  title = {{{AttriBoT}}: {{A Bag}} of {{Tricks}} for {{Efficiently Approximating Leave-One-Out Context Attribution}}},
  shorttitle = {{{AttriBoT}}},
  author = {Liu, Fengyuan and Kandpal, Nikhil and Raffel, Colin},
  year = {2025},
  number = {arXiv:2411.15102},
  eprint = {2411.15102},
  url = {https://arxiv.org/abs/2411.15102}
}

@misc{Beechey:2023,
  title = {Explaining {{Reinforcement Learning}} with {{Shapley Values}}},
  author = {Beechey, Daniel and Smith, Thomas M. S. and Simsek, Ozgur},
  date = {2023-06-09},
  year = {2023},
  eprint = {2306.05810},
  eprinttype = {arXiv},
  eprintclass = {cs},
  url = {http://arxiv.org/abs/2306.05810},
  pubstate = {prepublished}
}

@article{Farquhar:2024,
  title = {Detecting Hallucinations in Large Language Models Using Semantic Entropy},
  author = {Farquhar, Sebastian and Kossen, Jannik and Kuhn, Lorenz and Gal, Yarin},
  year = {2024},
  journal = {Nature},
  volume = {630},
  number = {8017},
  pages = {625--630}
}

@inproceedings{Kuhn:2023,
  title = {Semantic Uncertainty: {{Linguistic}} Invariances for Uncertainty Estimation in Natural Language Generation},
  booktitle = {The Eleventh International Conference on Learning Representations},
  author = {Kuhn, Lorenz and Gal, Yarin and Farquhar, Sebastian},
  year = {2023}
}

@inproceedings{He:2024,
  title = {{{OlympiadBench}}: A Challenging Benchmark for Promoting {{AGI}} with Olympiad-Level Bilingual Multimodal Scientific Problems},
  booktitle = {Proceedings of the 62nd Annual Meeting of the Association for Computational Linguistics (Volume 1: {{Long}} Papers)},
  author = {He, Chaoqun and Luo, Renjie and Bai, Yuzhuo and Hu, Shengding and Thai, Zhen and Shen, Junhao and Hu, Jinyi and Han, Xu and Huang, Yujie and Zhang, Yuxiang and Liu, Jie and Qi, Lei and Liu, Zhiyuan and Sun, Maosong},
  editor = {Ku, Lun-Wei and Martins, Andre and Srikumar, Vivek},
  year = {2024},
  pages = {3828--3850},
  address = {Bangkok, Thailand}
}

@misc{Shen:2025,
      title={Let's Verify Math Questions Step by Step}, 
      author={Chengyu Shen and Zhen Hao Wong and Runming He and Hao Liang and Meiyi Qiang and Zimo Meng and Zhengyang Zhao and Bohan Zeng and Zhengzhou Zhu and Bin Cui and Wentao Zhang},
      year={2025},
      eprint={2505.13903},
      archivePrefix={arXiv},
      primaryClass={cs.CL},
      url={https://arxiv.org/abs/2505.13903}, 
}

@misc{Numina:2024,
  title = {{{NuminaMath}}},
  author = {Li, Jia and Beeching, Edward and Tunstall, Lewis and Lipkin, Ben and Soletskyi, Roman and Huang, Shengyi Costa and Rasul, Kashif and Yu, Longhui and Jiang, Albert and Shen, Ziju and Qin, Zihan and Dong, Bin and Zhou, Li and Fleureau, Yann and Lample, Guillaume and Polu, Stanislas},
  year = {2024},
}

@misc{Khandoga:2026,
  title = {Beyond {{Uniform Credit}}: {{Causal Credit Assignment}} for {{Policy Optimization}}},
  shorttitle = {Beyond {{Uniform Credit}}},
  author = {Khandoga, Mykola and Yuan, Rui and Sankarapu, Vinay Kumar},
  year = {2026},
  url = {http://arxiv.org/abs/2602.09331},
  number = {arXiv:2602.09331},
  eprint = {2602.09331}
}

@inproceedings{Ielanskyi:2026,
  title = {Addressing Pitfalls in the Evaluation of Uncertainty Estimation Methods for Natural Language Generation},
  booktitle = {The Fourteenth International Conference on Learning Representations},
  author = {Ielanskyi, Mykyta and Schweighofer, Kajetan and Aichberger, Lukas and Hochreiter, Sepp},
  year = {2026},
  url = {https://openreview.net/forum?id=OxWnOV5q8w}
}

@misc{vonWerra:2020,
  title   = {{TRL: Transformers Reinforcement Learning}},
  author  = {von Werra, Leandro and Belkada, Younes and Tunstall, Lewis and Beeching, Edward and Thrush, Tristan and Lambert, Nathan and Huang, Shengyi and Rasul, Kashif and Gallouédec, Quentin},
  license = {Apache-2.0},
  url     = {https://github.com/huggingface/trl},
  year    = {2020}
}
	\bibliographystyle{plainnat}
	
	\newpage
	\appendix
	
	\section{Datasets and Models}
	
	\subsection{Datasets Used}
	\subsubsection{Dataset derived from \texttt{open-rs}}
	The models were trained on \texttt{open-rr} dataset.
	\texttt{open-rr} dataset is derived from \texttt{open-rs} dataset \citep{Dang:2025}.
	Despite the dataset having been compiled from other math datasets, most of which are supposed to be curated to any extent, our manual inspection has determined that it contained an error rate of $15$. Almost all of the relatively trivial-looking questions designated as 'Hard' and having no step-by-step solution were found to have incorrect labels.
	This could additionally explain why these questions were difficult for the model to learn: the reference solutions for them were incorrect!
	We have filtered the dataset further from $7,000$ examples to include only the examples containing the solution path, as well as on the length of the reference solution.
	
	\subsubsection{Evaluation Datasets}
	
	We used the following evaluation datasets: AIME24/25 \citep{aime24}, AMC, MATH-500 \citep{Hendrycks:2021a}, Minerva \citep{Lewkowycz:2022} and OlympiadBench \citep{He:2024}.
	The verification was done with \citep{Shen:2025}.
	The maximum generation length was chosen to be $4096$ for the experiments with 1.7B models and $25k$ tokens for the \texttt{Qwen3-4B} based experiments.
	
	\subsection{Models Used}
	
	We have made extensive use of Qwen2.5 distilled to Deepseek R1 traces \citep{DeepSeek-AI:2025} and Qwen3 \citep{Yang:2025} families of models.
	We avoided MoE models due to the increased complexity of training dynamics with them. 
	
	\section{Additional Figures}
	
	\begin{figure}[h]
		\centering
		\includegraphics[width=0.8\linewidth]{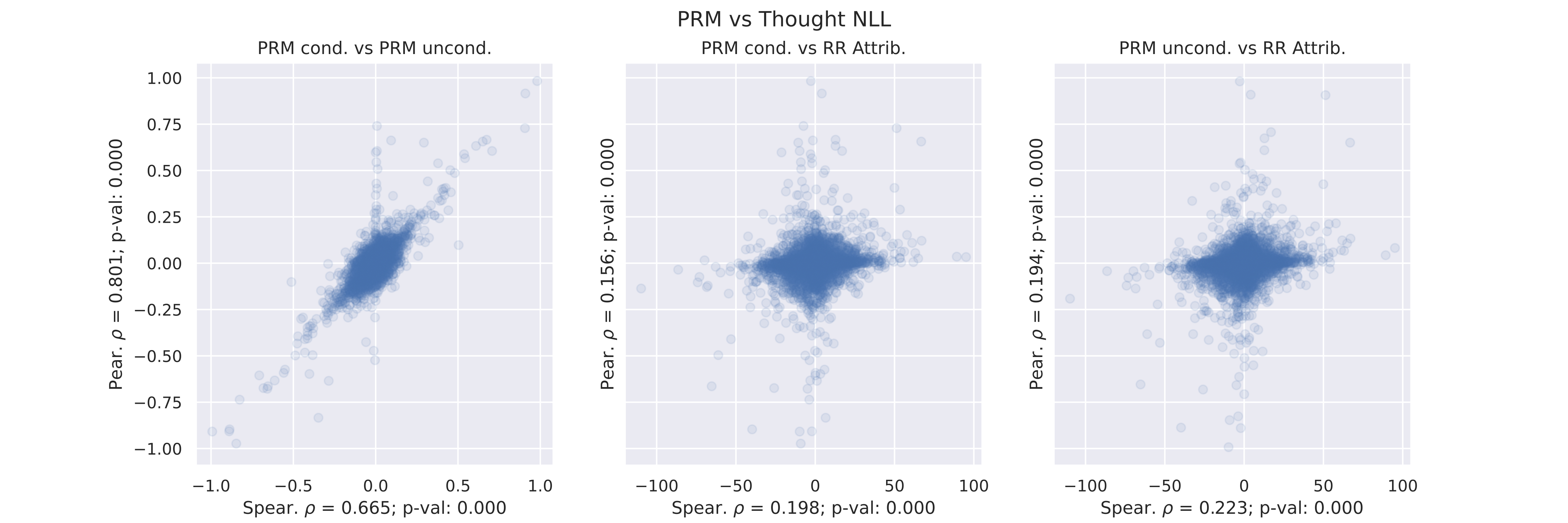}
		\includegraphics[width=0.8\linewidth]{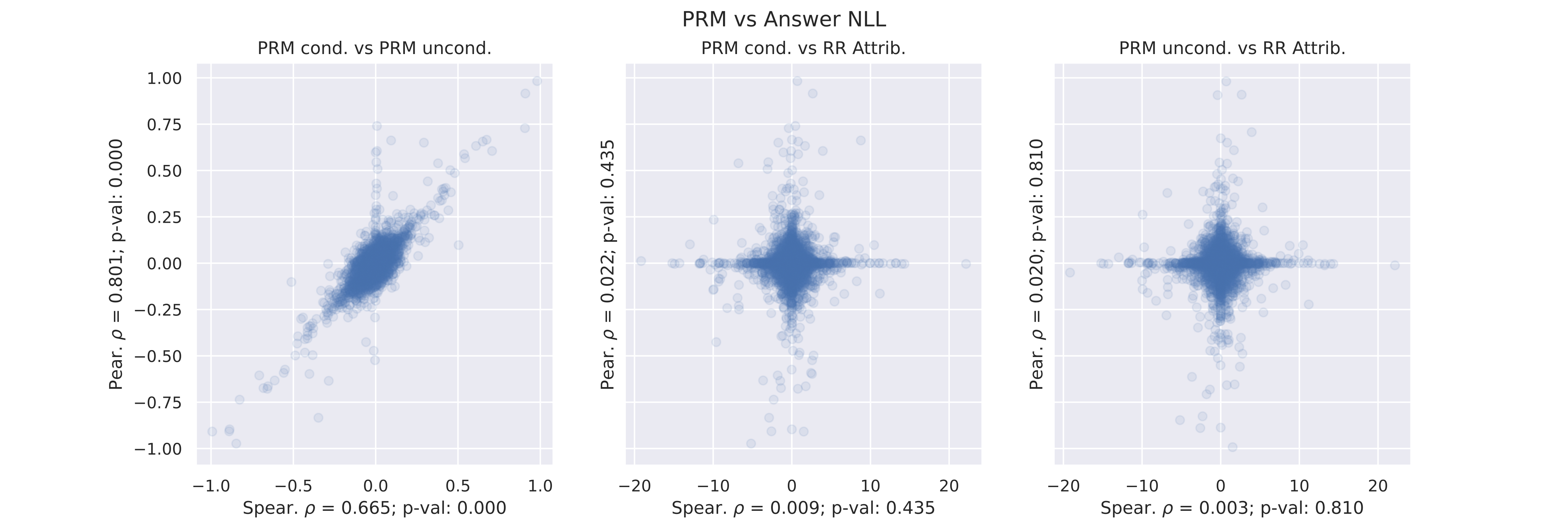}
		\caption{
			Correlation of PRM prediction to the different components of \methodname{} attribution. 
			PRM models seem to be more correlated to NLL changes of the thought NLL than to the change of NLL of the reference answer even when conditioned on it.
		}
		\label{appx:fig:prm_cor_to_components}
	\end{figure}
	
	\begin{figure}
		\centering
		\includegraphics[width=0.9\linewidth]{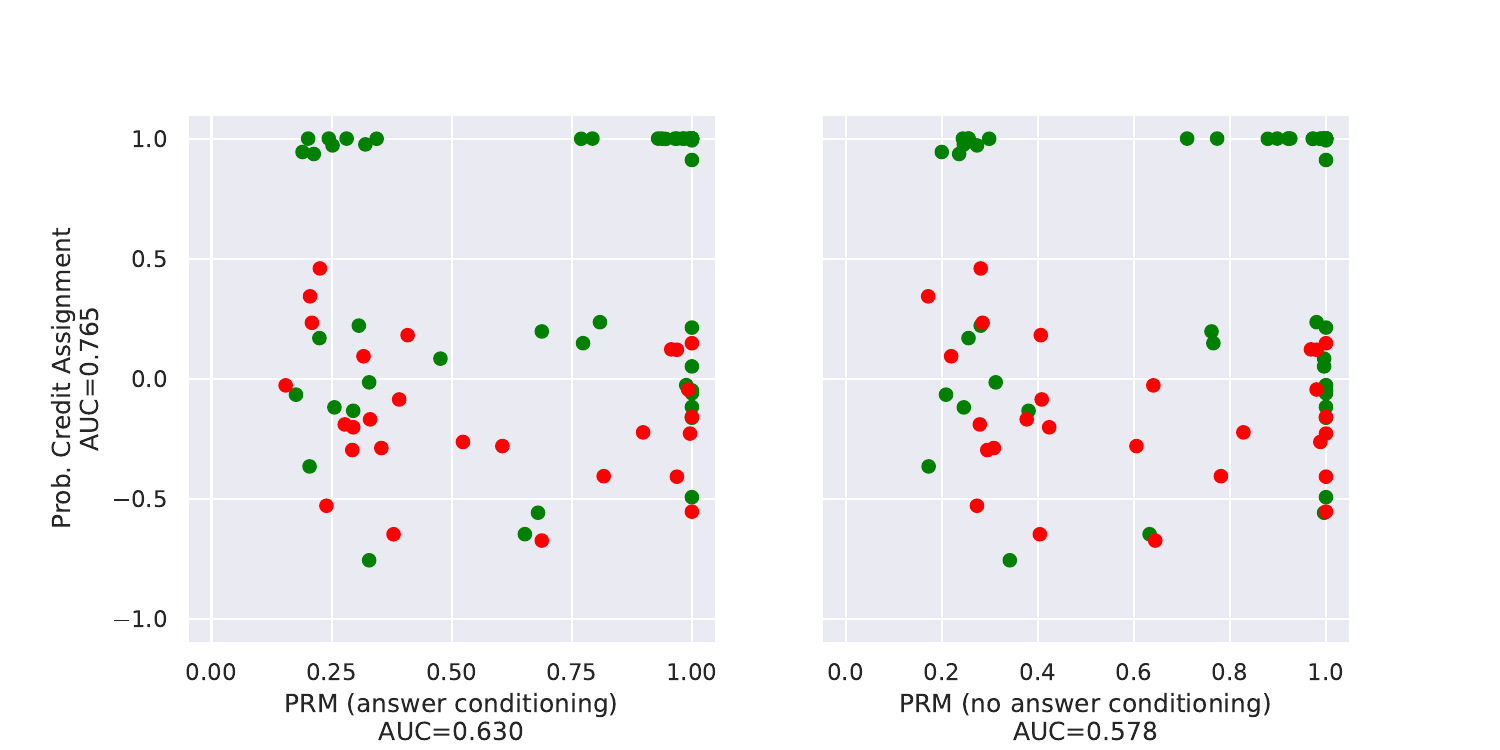}
		\caption{Correlation between the PRM and our credit assignment approach on $100$ examples from \texttt{open-rs} dataset.
			Green indicates that the ultimate answer was correct according to the verifiers, while red points are incorrect. 
			The x and y axes are the attribution by the PRM and our method, accordingly.
			The CoT traces were generated using \texttt{Deepseek-R1-Qwen2.5-7B-Distill} model while the PRM was computed using \texttt{Qwen2.5-Math-PRM-7B}. 
			Our attribution method results in better correlation to the ground truth than the PRM approach, even when the PRM is conditioned on the correct answer. 
		}
		\label{fig:prm_outcome_correlation}
	\end{figure}
	
	\begin{figure}[b!]
		\centering
		\includegraphics[width=0.45\linewidth]{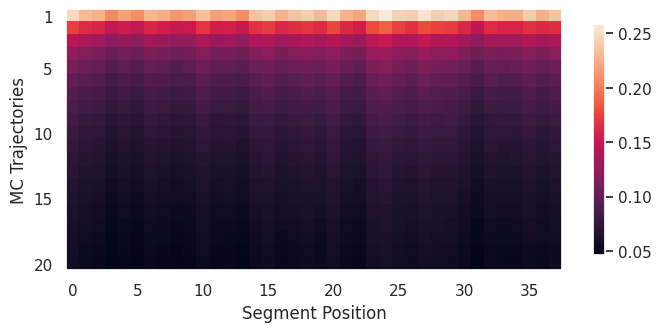}
		\caption{
			Standard Deviation of the MC value estimator. 
			The values were obtained by bootstrapping the original values from $20$ completions.
			Vertical axis is the number of MC samples - the number of rollouts selected for value estimation. 
			The horizontal axis is the number of exit points where the value was evaluated. 
		}
		\label{fig:mcts_sd_bootstrap}
	\end{figure}
	
	\section{Additional Equations}
	
	\subsection{Sequence Return Estimation}
	
	The RLPR and RLVR estimators are as follows:
	\begin{align}
	R^{VR}_{\Bu} & \approx \frac{1}{n} \sum_{i=0}^{n} \zeta(\By_{i}); \quad \By_{i} \drawnfrom p(. \mid \Bu, \Bx, \Bw)
	\label{appx:eq:ry_with_verifiers_mc_estimate} \\
	R^{PR}_{\Bu} & \approx \sum_{i=0}^{n} \zeta(\By_{i}) \; p(\By_{i} \mid \Bu, \Bx, \Bw) ; \quad \By_{i} \drawnfrom \{\By^\star\}^{n} \label{appx:eq:ry_probability_reward_importance_sampling}
	\end{align}
	
	Theorem : if $\By^\star$ contains all $\By$ such that $\zeta(\By_{i}) > 0$ then Eq.~\ref{appx:eq:ry_with_verifiers_mc_estimate} and Eq.~\ref{appx:eq:ry_probability_reward_importance_sampling} converge to the same $R_{\By}$. 
	
	This can be an obstacle for the practical application of RLPR when, e.g., the number of solutions is large or unbounded.
	
	\subsection{Transformations of Bellman Equation for CoT MDP}\label{appx:sec:equations_bellman}
	
	Value function decomposition for CoT MDP:
	\begin{align}
	v_{\Bw} (\Bs_{t}) = & \; \EXP[G_t \mid S_t = \Bs_{t}] \nonumber \\ 
	= & \; \EXP[R_{t+1} + \gamma G_{t+1} \mid S_t = \Bs_{t}] \nonumber \\ 
	= & \; \sum_{a} \pi(a \mid \Bs_t) \sum_{\Bs_{t+1}, r} p(\Bs_{t+1}, r \mid \Bs_{t}, a) \left[r + \gamma v_{\pi}(\Bs_{t+1})\right] \nonumber \\ 
	= & \sum_{a} \pi(a \mid \Bs_t) \sum_{\Bs_{t+1}, r} p(\Bs_{t+1}, r \mid \Bs_{t}, a) \ r + \sum_{a} \pi(a \mid \Bs_t) \sum_{\Bs_{t+1}, r} p(\Bs_{t+1}, r \mid \Bs_{t}, a) \ \gamma \ v_{\pi}(\Bs_{t+1}) \nonumber \\ 
	= & \sum_{\By} p(\By \mid \Bs, \Bw) \; r(\By) + \sum_{\Bu_t} p(\Bu_t \mid \Bs_{t}, \Bw) \sum_{\Bu_n} p(\Bs_{t+1} \mid \Bs_{t}, \Bu_t) \; \gamma \; v_{\pi}(\Bs_{t+1}) \nonumber \\
	= & \sum_{a \in \cY} p(a \mid \Bs_t, \Bw) \; r(\By) + \sum_{a \in \cU} p(a \mid \Bs_{t}, \Bw) \ \gamma \ v_{\Bw}(\Bs_{t+1})
	\end{align} \label{appx:eq:cot-mdp-bellman}
	
	This uses the return at time step $t$, defined as $G_t = \sum^{\infty}_{k=0} \gamma^{k} R_{t+k+1}$.
	
	Alternatively, we can reformulate the Bellman equation as estimation of a quantity under the Bayesian posterior of the reasoning language model $\Bw$:
	
	\begin{align} \label{appx:eq:bellman_to_bayesian_mdp}
	v_{\Bw} (\Bs_{t}) \; = & \; \EXP[G_t \mid S_t = \Bs_{t}] \\ 
	= & \; \EXP[R_{t+1} + \gamma G_{t+1} \mid S_t = \Bs_{t}] \nonumber \\ 
	= & \; \sum_{a} \pi(a \mid \Bs_t) \sum_{\Bs_{t+1}, r} p(\Bs_{t+1}, r \mid \Bs_{t}, a) \left[r + \gamma v_{\pi}(\Bs_{t+1})\right] \nonumber
	\end{align}
	From this point onward, given that (a) the dynamics function is trivial and (b) that $\gamma$ is set to $1.$ we can continue as follows:
	\begin{align}\label{appx:eq:cot-mdp-bellman-p2}
	= & \; \EXP_{\By, \Bu \drawnfrom p(\By, \Bu \mid \Bx, \Bw)} \bigg[ \zeta(\By) \bigg] \nonumber \\
	= & \; \sum^{\BY, \BU} p(\By, \Bu \mid \Bx, \Bw) \cdot \zeta(\By) \nonumber \\
	= & \; \sum^{\BY, \BU} p(\By \mid \Bu, \Bx, \Bw) p(\Bu \mid \Bx, \Bw) \cdot \zeta(\By) 
	\end{align} 
	
	\subsection{CoT Optimal Reward Redistribution Identity}
	Detailed derivation of the optimal reward redistribution for CoT generation:
	\begin{align}
	\EXP[R_{t+1}  \mid \Bs_{t-1}, a_{t-1}, \Bs_{t}, a_{t}] \; = & \; q^{\Bw} (\Bs_{t}, a_{t}) - q^{\Bw} (\Bs_{t-1}, a_{t-1}) \nonumber \\
	= & \; v^{\Bw}(\Bs_{t+1}) - v^{\Bw}(\Bs_{t}) \label{appx:eq:cot-mdp-value-diff} \\
	= & \;  \EXP_{\By \drawnfrom p(\By \mid \Bs_{t+1}, \Bw)} [ r(\By) ] + \EXP_{u_{t+1} \drawnfrom p(u_t \mid \Bs_{t+1}, \Bw)} [ v_{\pi}(\Bs_{t+2}) ] - \nonumber \\ 
	& \; - \EXP_{\By \drawnfrom p(\By \mid \Bs_{t}, \Bw)} [ r(\By) ] - \EXP_{\Bu_{t} \drawnfrom p(\Bu_t \mid \Bs_{t}, \Bw)} [ v_{\pi}(\Bs_{t+1}) ] \nonumber \\
	= & \; \left[ \EXP_{\By \drawnfrom p(\By \mid \Bs_{t+1}, \Bw)} [ r(\By) ] - \EXP_{\By \drawnfrom p(\By \mid \Bs_{t}, \Bw)} [ r(\By) ] \right] + \label{appx:eq:cot-mdp-goes-answer-part} \\
	& \; + \left[  \EXP_{\Bu_{t+1} \drawnfrom p(\Bu_{t+1} \mid \Bs_{t+1}, \Bw)} [ v^{\Bw}(\Bs_{t+2}) ] - \EXP_{\Bu_{t} \drawnfrom p(\Bu_t \mid \Bs_{t}, \Bw)} [ v^{\Bw}(\Bs_{t+1}) ] \right] \label{appx:eq:cot-mdp-goes-cot-part}
	\end{align}
	
	\subsection{Other Equations}
	
	General policy gradient \citep{Shao:2024}:
	\begin{align}
	\nabla_\theta \cJ_{\cA} = \EXP \underbrace{[(q,o) \drawnfrom \cD]}_{\text{Data Source}} \left( \frac{1}{|o|} \sum_{t=0}^{|o|} \underbrace{GC_{\cA} (q,o,t,\pi_{rf})}_{\text{Gradient Coefficient}} \underbrace{\nabla_\theta \log \pi_\theta (o_t \mid q, o_{<t})}_{\text{Token-Wise Gradient}} \right)
	\label{appx:eq:deepseek-math-objective-generalization}
	\end{align}
	
	\subsection{Value Estimator}\label{appx:sec:value_estimator_bias}
	
	Deriving the bias of the $\hat{v}^{our}(\Bs_t)$:
	\begin{align}
	\text{Bias} \left[ v_{\Bw}(\Bs_{t}),  \hat{v}^{our}_{\Bw} (\Bs_{t})\right] \; = & \; \EXP[\hat{v}^{our}_{\Bw} (\Bs_{t})] - v_{\Bw}(\Bs_{t}) \nonumber \\
	= & \; \frac{1}{N} \sum_{\By, \Bu \drawnfrom q} p(\By \mid \Bu, \Bx, \Bw) p(\Bu \mid \Bx, \Bw) \cdot \zeta(\By) \nonumber \\
	& - \frac{1}{Z} \sum_{\By \in \cY, \Bu \in \cU} p(\By \mid \Bu, \Bx, \Bw) p(\Bu \mid \Bx, \Bw) \cdot \zeta(\By) \label{appx:eq:bias:true_val} \\
	= & - \sum_{\By \in \cY \setminus \cY^\star, \Bu \in \cU \setminus \cU^\star} \frac{Z-N}{NZ} p(\By \mid \Bu, \Bx, \Bw) p(\Bu \mid \Bx, \Bw) \cdot \zeta(\By) \nonumber \\
	\approx & - \sum_{\By \in \cY \setminus \cY^\star, \Bu \in \cU \setminus \cU^\star} \frac{1}{N} p(\By \mid \Bu, \Bx, \Bw) p(\Bu \mid \Bx, \Bw) \cdot \zeta(\By)
	\end{align}
	For these transformations, we use the true value of the value function in Eq.~\ref{appx:eq:bias:true_val}, which is summed over the space of all possible final outputs and solution paths. 
	The penultimate transition is made by moving the factor inside the sum and essentially means that for every pair $\By^\star$ and $\Bu^\star$ we sum the integrand over all sequences that are not the specific sequence from the reference pool.
	The final transition is made by using the fact that $N \ll Z$.
	This bias is non-positive.
	It is zero if our $q$ covers the entirety of $\cY, \cU$. 
	
	\section{Details on Experimental Settings}\label{appx:sec:experiment_details}
	
	\subsection{MC Sampling Variance Experiment}\label{appx:sec:exp:mcts_variance}
	
	We have computed a single rollout for each question in the MATH-500 dataset. 
	The maximum completion length was $8196$.
	The generation and evaluation were performed with the Qwen3-4B-Thinking-2507 model. 
	Segmentation was performed using the hybrid method with a maximum number of segments set to $40$.
	$20$ completions were generated at every exit point, each keeping the maximum completion lengths constraint.
	This procedure took approximately $80$ GPU-hours on RTX 5000 PRO. 
	$100$ samples were subsampled to bootstrap the estimator. 
	
	\subsection{Attribution and Credit Assignment Method Correlation}\label{appx:sec:exp:attribution_credit_assignment}
	
	The same rollouts and model as in the previous subsection were used. 
	In addition, we have computed gradient, LOO and \methodname{} values for the segments.
	
	The LOO values were computed as pointwise mutual information between the segment and all of the subsequent segments under the model's predictive distribution.
	The gradient attribution was computed by computing the gradient of logprobs of the selected completion tokens with respect to input embeddings.
	To get per segment value out of that, the first-order norm of these gradients was taken and averaged per segment. 
	This is reminiscent of the gradient propagation occurring during the model update.
	
	The loo value computation took $0.5$ GPU-hours, while \methodname{} and gradient took about $0.3$ GPU-hours.
	The considerably more computationally expensive MC sampling was used as a proxy for ground truth value segment-wise advantage estimates. 
	
	\subsection{On Policy Model Training}
	
	\paragraph{Small Scale Experiments with \texttt{open-rs} dataset.}
	
	The setting and hyperparameters from \citep{Dang:2025} were used. 
	Hybrid segmentation was performed with the maximum segment number of $120$. 
	Completion length of $1024$ were used in training.
	
	For validation during training, we used avg@4 strategy with 4 datasets: Minerva (first 50), AIME24 (full), AIME25 (full) and MATH-500 (first 50). 
	The generation length was limited to $4096$ tokens for performance reasons.
	Answer extraction and validation were performed using the \texttt{math-verify} package. 
	This experiment is estimated to have consumed on the order of $500$ GPU-hours.
	
	\paragraph{Experiments with \texttt{Qwen3-4B-Instruct-2507}}
	
	Numina-CoT \citep{Numina:2024} dataset was used to train reasoning starting with \texttt{Qwen3-4B-Instruct-2507} \citep{Yang:2025}.
	The shared hyperparameters of GRPO and \methodname{} run were identical, with the key difference being the learning rate that was $1.e-6$ for GRPO and $5.e-7$ for \methodname{}.
	Note, that the learning rate was kept at the given value for the GRPO as it was found to perform worse with $5.e-7$. 
	In other words, the LR is individually tailored for each of the two algorithm. 
	The group size was set to $4$, gradient accumulation to $1$, total step size to $500$ with cosine LR scaling.
	
	For validation we used avg@$8$ strategy with 5 datasets: AIME24, AIME25, AIME26, Minerva-Math and MATH-500. 
	The generation length was limited to $25K$, similar to training, to assess the method performance at larger scale.
	Answer extraction and validation were performed using the \texttt{math-verify} package. 
	This experiment is estimated to have consume on the order of $1000$ GPU-hours.
	
\end{document}